\documentclass[nohyperref]{article}

\usepackage{microtype}
\usepackage{graphicx}
\usepackage{booktabs} 

\usepackage{hyperref}

\usepackage[accepted]{icml2022}

\usepackage{amsmath}
\usepackage{amssymb}
\usepackage{mathtools}
\usepackage{amsthm}

\usepackage[utf8]{inputenc} 
\usepackage[T1]{fontenc}    

\usepackage{booktabs}       
\usepackage{amsfonts}       
\usepackage{nicefrac}       
\usepackage{microtype}      

\usepackage{xcolor}    
\usepackage{lipsum}
\usepackage{graphicx}
\usepackage{algorithm}
\usepackage{algorithmic}
\usepackage{comment}

\usepackage{graphicx}
\usepackage{float}
\usepackage[capitalize,noabbrev]{cleveref}
\usepackage{multicol}
\usepackage{subfig}

\theoremstyle{plain}

\theoremstyle{definition}

\theoremstyle{remark}

\DeclareMathOperator*{\argmax}{arg\,max}

\newcommand{\norm}[1]{\left\lVert#1\right\rVert}

\usepackage[disable,textsize=tiny]{todonotes}
\usepackage[textsize=tiny]{todonotes}

\icmltitlerunning{Submission and Formatting Instructions for ICML 2022}

\begin{document}

\twocolumn[
\icmltitle{Robust Reinforcement Learning with  Distributional Risk-averse formulation}




\begin{icmlauthorlist}
\icmlauthor{Pierre Clavier}{yyy,comp}
\icmlauthor{Stéphanie Allassonnière}{comp}
\icmlauthor{Erwan Le Pennec}{yyy}


\end{icmlauthorlist}

\icmlaffiliation{yyy}{Centre de Mathématiques Appliquées, Ecole polytechnique, France}
\icmlaffiliation{comp}{INRIA HeKA, INSERM, Sorbonne
Université, Université Paris Cité, France}

\icmlcorrespondingauthor{Pierre Clavier }{pierre.clavier@polytechnique.edu}

\icmlkeywords{Machine Learning, ICML}

\vskip 0.3in
]



\printAffiliationsAndNotice{} 

\begin{abstract}

 Robust Reinforcement Learning tries to make predictions more robust to changes in the dynamics or rewards of the system. This problem is particularly important when the dynamics and rewards of the environment are estimated from the data. In this paper, we approximate the Robust Reinforcement Learning constrained with a $\Phi$-divergence using an approximate Risk-Averse formulation. We show that the classical Reinforcement Learning formulation can be robustified using standard deviation penalization of the objective. Two algorithms based on Distributional Reinforcement Learning, one for discrete and one for continuous action spaces are proposed and tested in a classical Gym environment to demonstrate the robustness of the algorithms.
\end{abstract}

\section{Introduction}
\label{submission}

The classical Reinforcement Learning (RL) problem using Markov Decision Processes (MDPs) modelization gives a practical framework for solving sequential decision problems under uncertainty of the environment. However, for real-world applications, the final chosen policy can be sometimes very sensitive to sampling errors, inaccuracy of the model parameters, and definition of the reward. For discrete state-action space, Robust MDPs is treated by \citet{Yang2017,petrik2019beyond,grand2020first,grand2020scalable} or \citep{behzadian2021fast} among others. Here we focus on \textit{more general continuous state space $\mathcal{S}$} with discrete or continuous action space $\mathcal{A}$ and with constraints defined using $\Phi$-divergence. Robust RL \citep{morimoto2005robust} with continuous action space focuses on robustness in the dynamics of the system (changes of $P$)  and has been studied in 
 \citet{abdullah2019wasserstein,singh2020improving,urpi2021risk,eysenbach2021maximum} among others.  \citet{eysenbach2021maximum} tackles the problem of both reward and transitions using Max Entropy RL whereas the problem of robustness in action noise perturbation is presented in \citet{tessler2019action}. Here we tackle the problem of Robustness \textit{thought dynamics of the system.}
 
In this paper, we show that it is possible to approximate a Robust Distributional Reinforcement Learning heuristic with $\Phi$-divergence constraints into a Risk-averse formulation, using a formulation based on mean-standard deviation optimization. Moreover, we focus on the idea that generalization, regularization, and robustness are strongly linked together as  \citet{husain2021regularized,derman2020distributional,derman2021twice,ying2021towards,brekelmans2022your} noticed in the MDPs or RL framework. The contribution of the work is the following: we motivate the use of standard deviation penalization and derive two algorithms for discrete and continuous action space that are Robust to change in dynamics. These algorithms do not require lots of additional parameter tuning, only the Mean-Standard Deviation trade-off has to be chosen carefully. Moreover, we show that our formulation using Distributional Reinforcement Learning is robust to change dynamics on discrete and continuous action spaces from Mujoco suite.
\section{Notations}
 Considering a Markov Decision Process (MDP)  $(\mathcal{S}, \mathcal{A}, P, \gamma)$, where  $\mathcal{S}$ is the state space, $\mathcal{A}$ is the action space, $P\left(s^{\prime},r \mid s, a\right)$ is the reward and transition distribution from state $s$ to $s' $ taking action $a$ and  $\gamma \in(0,1)$ is the discount factor. Stochastic policy are denoted $\pi(a \mid$ $s): \mathcal{S} \rightarrow \Delta(\mathcal{A})$ and we consider the cases of action space either discrete our continuous.
 
 A rollout or trajectory using $\pi$ from state $s$ using initial action $a$ is defined as the the random sequence $\tau^{P,\pi\mid s,a}=\left(\left(s_{0}, a_{0}, r_{0}\right),\left(s_{1}, a_{1}, r_{1}\right), \ldots\right)$ with  $s_{0}=s, a_{0}=a, a_{t} \sim \pi\left(\cdot \mid s_{t}\right)$ and $(r_{t},s_{t+1}) \sim P\left(\cdot,\cdot \mid s_{t}, a_{t}\right) ;$ we denote the distribution over rollouts by $\mathbb{P(\tau)}$ with $ \mathbb{P}(\tau)=P_{0}\left(s_0\right) \prod_{t=0}^{T} P\left(s_{t+1}, r_t\mid s_t, a_t\right) \pi \left(a_t \mid s_t\right) d \tau $ and usually write $\tau \sim \mathbb{P}= (P,\pi)$. Moreover, considering the distribution of discounted cumulative return $Z^{P,\pi}(s,a)=R(\tau^{P,\pi\mid s,a})$ with $R(\tau)=\sum_{t=0}^{\infty} \gamma^{t} r_t$, the $Q$-function $Q^{P,\pi}: \mathcal{S} \times \mathcal{A} \rightarrow \mathbb{R}$ of $\pi$ is the  expected discounted cumulative return of the distribution defined as follows : $ Q^{P,\pi}(s, a):=\mathbb{E}[Z^{P,\pi}(s,a)]$. The classical initial goal of  $\mathrm{RL}$ also called risk-neutral RL, is to find the optimal policy $\pi^{*}$ where $Q^{P,\pi^{*}}(s, a) \geq Q^{P,\pi}(s, a)$  for all $\pi$ and $s \in \mathcal{S}, a \in \mathcal{A}$. Finally, the Bellman operator $\mathcal{T}^{\pi}$ and Bellman optimal operator $\mathcal{T}^{*}$ are defined as follow:
$\mathcal{T}^{\pi} Q(s, a):=r(s,a)+\gamma \mathbb{E}_{P, \pi}\left[Q\left(s^{\prime}, a^{\prime}\right)\right] $ and 
$\mathcal{T}^{*} Q(s, a):=r(s,a)+\gamma  \mathbb{E}_{P}\left[\max _{a^{\prime}}Q\left(s^{\prime}, a^{\prime}\right)\right]$.

 Applying either operator from an initial $Q^{0}$ converges to a fixed point $Q^{\pi}$ or $Q^{*}$ at a geometric rate as both operators are contractive. Simplifying the notation with regards to $s,a ,\pi$ and  $P$, we define the set of greedy policies w.r.t. $Q $ called $\displaystyle\mathcal{G}(Q)=\arg\max \limits _{\pi \in \Pi} \langle Q, \pi\rangle$. A classical approach to estimating an optimal policy is known as Approximate Modified Policy Iteration (AMPI) \cite{scherrer2015approximate}
$$
\left\{\begin{array}{l}
\pi_{k+1} \in \mathcal{G}\left(Q_{k}\right) \\
Q_{k+1}=\left(T^{\pi_{k+1}}\right)^{m} Q_{k}+\epsilon_{k+1}
\end{array},\right.
$$
which usually reduces to Approximate Value Iteration (AVI, $m=1$ ) and Approximate Policy Iteration $(\mathrm{API}, m=\infty)$ as special cases. The term $\epsilon_{k+1}$ accounts for errors made when applying the Bellman Operator in RL algorithms with stochastic approximation.

\section{Robust formulation in greedy step of API.}
In this section, we would like to find a policy that is robust to change of environment law $P$ as small variations of $P$ should not affect too much the new policy in the greedy step.  In our case we are not looking at classical greedy step $\displaystyle\pi'\in \mathcal{G}(Q)= \arg\max \limits _{\pi \in \Pi} \langle Q, \pi\rangle$ rather at the following  :$$\pi'\in \mathcal{G}(Q)=\argmax _{\pi \in \Pi} \langle \min _P Q^{(P,\pi)}, \pi\rangle.$$ This heuristic in the greedy step can also be justified by trying to avoid an overestimation of the Q functions present in the Deep RL algorithms. Using this formulation, we need to constrain the set of admissible transitions from state-action to the next state $P$ to get a solution to the problem. In general, without constraint, the problem is  NP-Hard, so it requires constraining the problem to specific distributions that are not too far from the original one using distance between distributions such as the Wasserstein metric \citep{abdullah2019wasserstein} or other specific distances where the problem can be simplified \citep{eysenbach2021maximum}. If a explicit form of  $ \min _P Q^{(P,\pi)}$ could be computed exactly for a given  divergence, it would lead to a simplification of this max-min optimization problem into a simple maximisation one.
 
In fact, simplification of the problem is possible using specific $\Phi$-divergence denoted $\mathcal{H}_{\Phi}$ to constrain the problem with $\Phi$ a closed convex function such that $\Phi: \mathbb{R} \rightarrow \mathbb{R} \cup\{+\infty\}$ and $\Phi(z) \geq \Phi(1)=0$ for all $z \in \mathcal{R}$ : $\mathcal{H}_{\Phi}\left(\mathbb{Q} \mid \mathbb{P}\right) = 
\sum_{i: p_{i}>0} p_{i} \Phi\left(\frac{q_{i}}{p_{i}}\right) 
$ with  $\sum_{i: p_{i}>0} q_{i}=1 $ and $ q_{i} \geq 0 $. 
 This constraint requires $q_i=0$ if $p_i=0$ which makes the measure $\mathbb{Q}$  absolutely continuous with respect to $\mathbb{P}$.The $\chi^2 $-divergence are a particular case of $\Phi$-divergence with $\Phi(z)=(z-1)^2$. For trajectories sampled from distribution $P_0$ and looking at distribution $P$ closed to $P_0$ with regards to  the $\chi^2 $-divergence, the minimisation problem reduces to : 
\begin{equation}
    \label{eq_cauchy}
    \min  _{P \in D_{\chi^2}(P\| P_0)\leq \alpha} Q^{(P,\pi)}= Q^{(P_0,\pi)}-\alpha\mathbb{V}_{P_0}[Z]^{\frac{1}{2}}.
\end{equation}
The proof can be found in annex A for $\alpha$ such that   $\alpha\leq \mathbb{V}_{P_0}[R(\tau)]/\norm{\tilde{R}}_{\infty}^2\leq 1$ with $\tilde{R}(\tau)=R(\tau)-\mathbb{E}_{\tau \sim P_0}[R(\tau)]$ the centered return and $\mathbb{V}_{P_0}[Z]$ the variance of returns.  For $\alpha> \mathbb{V}_{P_0}[R(\tau)]/\norm{\tilde{R}}_{\infty}^2$,  the equality becomes an inequality but we still optimize a lower bound of our initial problem. Defining a new greedy step which is penalized by the standard deviation : 
$$\pi'\in \mathcal{G}_\alpha(Q)=\argmax _{\pi \in \Pi } \langle Q^{(P_0,\pi)}-\alpha\mathbb{V}_{P_0}[Z]^{\frac{1}{2}} , \pi\rangle$$
we now look at the current AMPI to improve robustness :$$
\left\{\begin{array}{l}
\pi_{k+1} \in \mathcal{G}_\alpha \left(Q_{k}\right) \\
Q_{k+1}=\left(T^{\pi_{k+1}}\right)^{m} Q_{k}+\epsilon_{k+1}
\end{array},\right.
$$
 This idea is very close to Risk-averse formulation in RL (i.e minimizing risk measure and not only the mean of rewards)  but here the idea is to approximate a robustness problem in RL. To do so, the standard deviation of the distribution of the returns must be estimated. Many ways are possible but we favour distributional RL \citep{Bellemare2017,Dabney2017,Dabney2018} which achieve very good performances in many RL applications. Estimating quantiles of the distribution of return, we can simply estimate standard deviation using classical estimator of the standard deviation given the quantiles over an uniform grid$ \{q_i(s,a)\}_{1\leq i \leq n}, :$
$\mathbb{V}[Z(s,a)]^{\frac{1}{2}}=\sigma^{}(s, a)=\sqrt{\sum_{i=1}^{n}\left(q_{i}^{}(s, a)-\bar{q}^{}(s, a)\right)^{2}}$
where $\bar{q}$ is the classical estimator of the mean. A different interpretation of this formulation could be that by taking actions with less variance, we construct a confidence interval with the standard deviation of the distribution :
$$Z^{\pi}(s, a) \stackrel{d}{=} \bar{Z}(s, a)-\alpha \sigma(s, a).$$
This idea is present in classical UCB algorithms  \citep{auer2002using} or pessimism/optimism Deep RL. Here we construct a confidence interval using the distribution of the return and not different estimates of the $Q$ function such as in \citet{moskovitz2021tactical,bai2022pessimistic}. In the next section, we derive two algorithms, one for discrete action space and one for continuous action space using this idea. A very interesting way of doing robust Learning is by doing Max entropy RL such as in the SAC algorithm. In \citet{eysenbach2021maximum}, a demonstration that SAC is a surrogate of Robust RL is demonstrated formally and numerically and we will compare our algorithm to this method.

\section{Distributional RL}
Distributional RL aims at approximating the return random variable $Z^{\pi}(s, a):=$ $\sum_{t=0}^{\infty} \gamma^{t} \mathcal{R}\left(s_{t}, a_{t}\right)$ with $s_{0}=s, a_{0}=a$ , $s_{t+1} \sim$ $\mathcal{P}\left(\cdot \mid s_{t}, a_{t}\right), a_{t} \sim \pi\left(\cdot \mid s_{t}\right)$and $\mathcal{R}$ that we explicitly treat as a random
variable. The classical RL framework  approximate the expectation of the return or the Q-function, $Q^{\pi}(s, a):=\mathbb{E}\left[Z^{\pi}(s, a)\right] .$ Many algorithms and distributional representation of the critic exits \citep{Bellemare2017,Dabney2017,Dabney2018} but here we focus on QR-DQN \cite{Dabney2017} that approximates the distribution of returs $Z^{\pi}(s, a)$ with $Z_{\psi}(s, a):=\frac{1}{M} \sum_{m=1}^{M} \delta\left(\theta_{\psi}^{m}(s, a)\right)$, a mixture of atoms-Dirac delta functions  located at $\theta_{\psi}^{1}(s, a), \ldots, \theta_{\psi}^{M}(s, a)$ given by a parametric model $\theta_{\psi}:$ $\mathcal{S} \times \mathcal{A} \rightarrow \mathbb{R}^{M}$. Parameters $\psi$ of a neural network are obtained by minimizing the average over the 1-Wasserstein distance between $Z_{\psi}$ and the temporal difference target distribution $\mathcal{T}_{\pi} Z_{\bar{\psi}}$, where $\mathcal{T}_{\pi}$ is the distributional Bellman operator defined in \citet{Bellemare2017}. The control version or optimal operator is denoted   $\mathcal{T} Z_{\bar{\psi}}$,
with $\mathcal{T}^{\pi} Z(s, a)= \mathcal{R}(s, a)+\gamma Z\left(s^{\prime}, a^{\prime}\right)$. According to \citet{Dabney2017}, the minimization of the 1-Wasserstein loss can be done by learning quantile locations for fractions $\tau_{m}=$ $\frac{2 m-1}{2 M}, m \in[1 . . M]$  via quantile regression loss, defined for a quantile fraction $\tau \in[0,1]$ as :
$$
\begin{aligned}
\mathcal{L}_{\mathrm{QR}}^{\tau}(\theta): &=\mathbb{E}_{\tilde{Z} \sim Z}\left[\rho_{\tau}(\tilde{Z}-\theta)\right]
\end{aligned}
$$
with $\rho_{\tau}(u) =u(\tau-\mathbb{I}(u<0)), \forall u \in \mathbb{R}$ . Finally, to obtain better gradients when $u$ is small, Huber quantile loss (or asymmetric Huber loss) can be used:
$$
\rho_{\tau}^{H}(u)=|\tau-\mathbb{I}(u<0)| \mathcal{L}_{H}^{1}(u),
$$
where $\mathcal{L}_{H}^{1}(u)$ is a classical Huber loss with parameter 1.
The quantile representation has the advantage of not fixing the support of the learned distribution and is used to represent the distribution of return in our algorithm for both discrete and continuous action space.
\section{Algorithm}
We use a distributional maximum entropy framework for continuous action space which is closed to the TQC algorithm \cite{kuznetsov2020controlling}.  This method uses an actor-critic framework with a distributional truncated critic to ovoid overestimation in the estimation with the max operator. This algorithm is based on a soft-policy iteration where we penalize the target $y_i(s,a)$ using the entropy of the distribution. More formally, to compute the target, the principle is to train $N$ approximate estimate $Z_{\psi_{1}}, \ldots Z_{\psi_{C}}$ of the distribution of returns $Z^{\pi}$ where $Z_{\psi_{c}}$ maps each $(s, a)$ to $Z_{\psi_{c}}(s, a):=\frac{1}{M} \sum_{m=1}^{M} \delta\left(\theta_{\psi_{n}}^{m}(s, a)\right),$
which is supported on atoms $\theta_{\psi_{c}}^{1}(s, a), \ldots, \theta_{\psi_{c}}^{M}(s, a)$.
Then approximations $Z_{\psi_{1}}, \ldots Z_{\psi_{N}}$ are trained on the temporal difference target distribution denoted $Y(s, a)$ constructed as follow. First atoms of trained distributions $Z_{\psi_{1}}\left(s^{\prime}, a^{\prime}\right), \ldots, Z_{\psi_{C}}\left(s^{\prime}, a^{\prime}\right)$ are pooled into 
$
\mathcal{Z}\left(s^{\prime}, a^{\prime}\right):=\left\{\theta_{\psi_{c}}^{m}\left(s^{\prime}, a^{\prime}\right) \mid c \in[1 . . C], m \in[1 . . M]\right\}
$. We denote elements of $\mathcal{Z}\left(s^{\prime}, a^{\prime}\right)$ sorted in ascending order by $z_{(i)}\left(s^{\prime}, a^{\prime}\right)$, with $i \in[1 . . M C]$.
Then we only keep the $k C$ smallest elements of $\mathcal{Z}\left(s^{\prime}, a^{\prime}\right)$. We remove outliers of distribution to avoir overestimation of the value function. Finally the atoms of the target distribution $
Y(s, a):=\frac{1}{k C} \sum_{i=1}^{k C} \delta\left(y_{i}(s, a)\right)
$ are computed according to a soft policy gradient method where we penalised with the $\log $ of the policy :
  \begin{equation}
    y_{i}(s, a):=r(s, a)+\gamma\left[z_{(i)}\left(s^{\prime}, a^{\prime}\right)-\eta \log \pi_{\phi}\left(a^{\prime} \mid s^{\prime}\right)\right].
\end{equation}
The 1-Wasserstein distance between each of $Z_{\psi_{n}}(s, a), n \in[1 . . N]$ and the temporal difference target distribution $Y(s, a)$ is minimized learning the locations for quantile fractions $\tau_{m}=\frac{2 m-1}{2 M}, m \in[1 . . M]$. Similarly, we minimize the loss :
 \begin{align*}
 \resizebox{\linewidth}{!}{ $ J_{Z}\left(\psi_{c}\right)=\mathbb{E}_{\mathcal{D,\pi}}\left[\frac{1}{M k C } \sum_{j=1}^{M} \sum_{i=1}^{k C} \rho_{\tau_{j}}^{H}\left(y_{i}(s, a)-\theta_{\psi_{c}}^{j}(s, a)\right) \right]$ }
\end{align*}

over the parameters $\psi_{n}$, for each critic. With this formulation, the learning of all quantiles $\theta_{\psi_{n}}^{m}(s, a)$ is  dependent on all atoms of the truncated mixture of target distributions. To optimize the actor, the following loss based on KL-divergence denoted $\mathrm{D}_{\mathrm{KL}}$ is used for soft policy improvement, : 
 \begin{align*}
 \resizebox{\linewidth}{!}{ $ J_{\pi,\alpha}(\phi)=\mathbb{E}_{ \mathcal{D}}\left[\mathrm{D}_{\mathrm{KL}}\left(\pi_{\phi}\left(\cdot \mid s\right) \| \frac{\exp \left(\frac{1}{\eta}\xi_\alpha (\theta_{\psi}\left(s, \cdot\right))\right)}{D}\right)\right]$ }
\end{align*}
 where $\eta$ can be seen as a temperature and needs to be tuned and D is a constant of normalisation. This expression simplify into :
\begin{equation*}
    J_{\pi,\alpha}(\phi)=\mathbb{E}_{\mathcal{D}, \pi}\left[\eta \log \pi_{\phi}(a \mid s)-\frac{1}{C } \sum_{ c=1}^{ C}\xi_\alpha( \theta_{\psi_{c}}(s, a))\right]
\end{equation*}
where $s \sim \mathcal{D}, a \sim \pi_{\phi}(\cdot \mid s)$.  Nontruncated estimate of the Q-value are used for policy optimization to avoid a double truncation, in fact the $Z$-functions already approximate truncated future distribution. Finally, $\eta$ is the entropy temperature coefficient and is dynamically adjusted by taking a gradient step with respect to the loss like in \citet{Haarnoja2018} :
$$
J(\eta)=\mathbb{E}_{\mathcal{D}, \pi_{\phi}}\left[\left(-\log \pi_{\phi}\left(a_{t} \mid s_{t}\right)-\mathcal{H}_{\eta}\right) \eta \right]
$$
at every time the $\pi_{\phi}$ changes. Temperature $\eta$ decreases if the policy entropy, $-\log \pi_{\phi}\left(a_{t} \mid s_{t}\right)$, is higher than $\mathcal{H}_{\eta}$ and increases otherwise. 
 The algorithm is summarized as follows :
\begin{algorithm}
\begin{algorithmic}
\caption{ TQC  with Standard Deviation penalisation}
\STATE\textbf{Initialize}  policy $\pi_{\phi}$, critics $Z_{\psi_{c}}, Z_{\bar{\psi}_{c}}$ for $c \in[1 . . C]$
\FOR{ each iteration }
\FOR{each step of the environment}
\STATE collect  $\left(s_{t}, a_{t}, r_{t}, s_{t+1}\right)$ with policy $\pi_{\phi}$ 
\STATE$\mathcal{D}  \leftarrow  \mathcal{D} \cup\left\{\left(s_{t}, a_{t}, r_{t}, s_{t+1}\right)\right\}$
\ENDFOR
\FOR{each gradient steps}
\STATE Sample  batch $(s,a,s',r)$ of $\mathcal{D} $

\STATE $ y_{i}(s, a)\leftarrow r(s, a)+\gamma\left[z_{(i)}\left(s^{\prime}, a^{\prime}\right)-\eta \log \pi_{\phi}\left(a^{\prime} \mid s^{\prime}\right)\right]$
\STATE$\eta \leftarrow \eta-\lambda_{\eta} \hat{\nabla}_{\eta} J_{}(\eta)$
\STATE$\phi \leftarrow \phi-\lambda_{\pi} \hat{\nabla}_{\phi} J_{\pi,\alpha}(\phi)$
\STATE $\psi_{c} \leftarrow \psi_{c}-\lambda_{Z} \hat{\nabla}_{\psi_{c}} J_{Z}\left(\psi_{n}\right), c \in[1 . . C]$
\STATE $\bar{\psi}_{c} \leftarrow \beta \psi_{c}+(1-\beta) \bar{\psi}_{c}, c \in[1 . . C]$

\ENDFOR
\ENDFOR

\textbf{return} policy $\pi_{\phi}$,$Z_{\psi_{c}}, c \in[1 \hdots C]$.
\end{algorithmic}
\end{algorithm}
Our algorithm is based SAC framework but with many distributional critics to improve the estimation of $Q$-functions while using mean-standard deviation objective in the policy loss to improve robustness.

\section{Experiments}
We try different experiments on continuous and discrete action space (see Annex B.2. for discrete case) to demonstrate the interest of our algorithms for robustness using  $\xi :Z \rightarrow \mathbb{E}[Z]-\alpha \mathbb{V}[Z]^{\frac{1}{2}}$ instead of the mean. For continuous action space, we compare our algorithm with SAC which achives state of the art in robust control \citep{eysenbach2021maximum} on the Mujoco environment such as Hopper-v3, Walker-v3, or HalfCheetah-v3. We use a version where the entropy coefficient is adjusted during learning for both SAC and our algorithm as it requires less parameter tuning. Moreover, we show the influence of a distributional critic without a mean-standard deviation greedy step using $\alpha =0$ to demonstrate the advantage of using a distributional critic against the classical SAC algorithm. We also compare our results to TQC algorithm which is in fact very close to SAC algorithm except the distributional critic. Finally, $\alpha$ the penalty is increased to show that for the tested environment, there exists a value of $\alpha$ such as prediction are more robust to change of dynamics. In these simulations, variations of dynamics are carried out by moving the relative mass which is an influential physical parameter in all environments. All algorithms are trained with a relative mass of 1 and then tested on new environment where the mass varies from $0.5$ to $2$. Two phenomena can be observed for the 3 environments.

\begin{figure}[ht]
     \centering
     \subfloat[Hopper-v3]{\includegraphics[width=0.45\columnwidth]{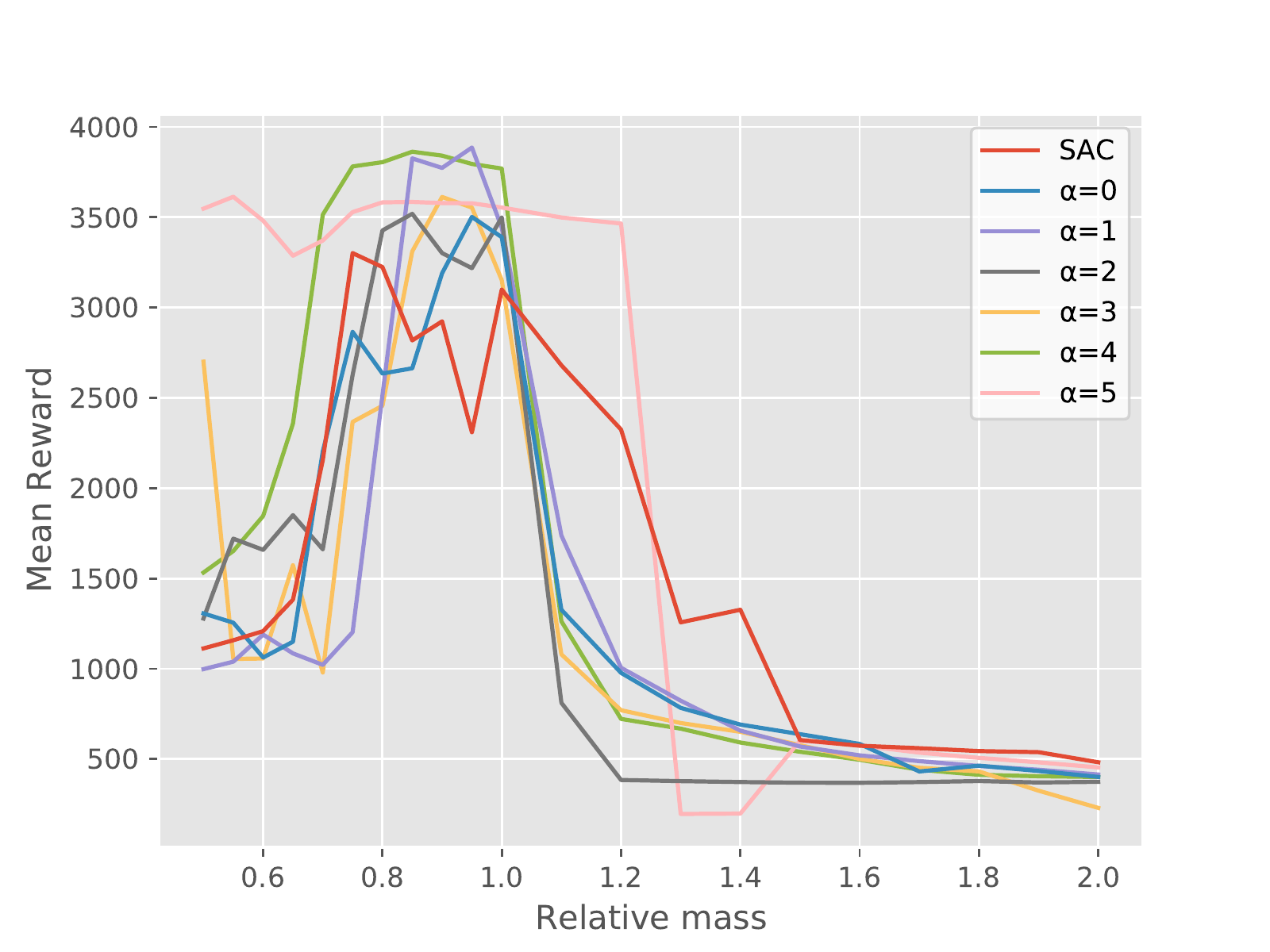}}
     \subfloat[Walker2d-v3]{ \includegraphics[width=0.45\columnwidth]{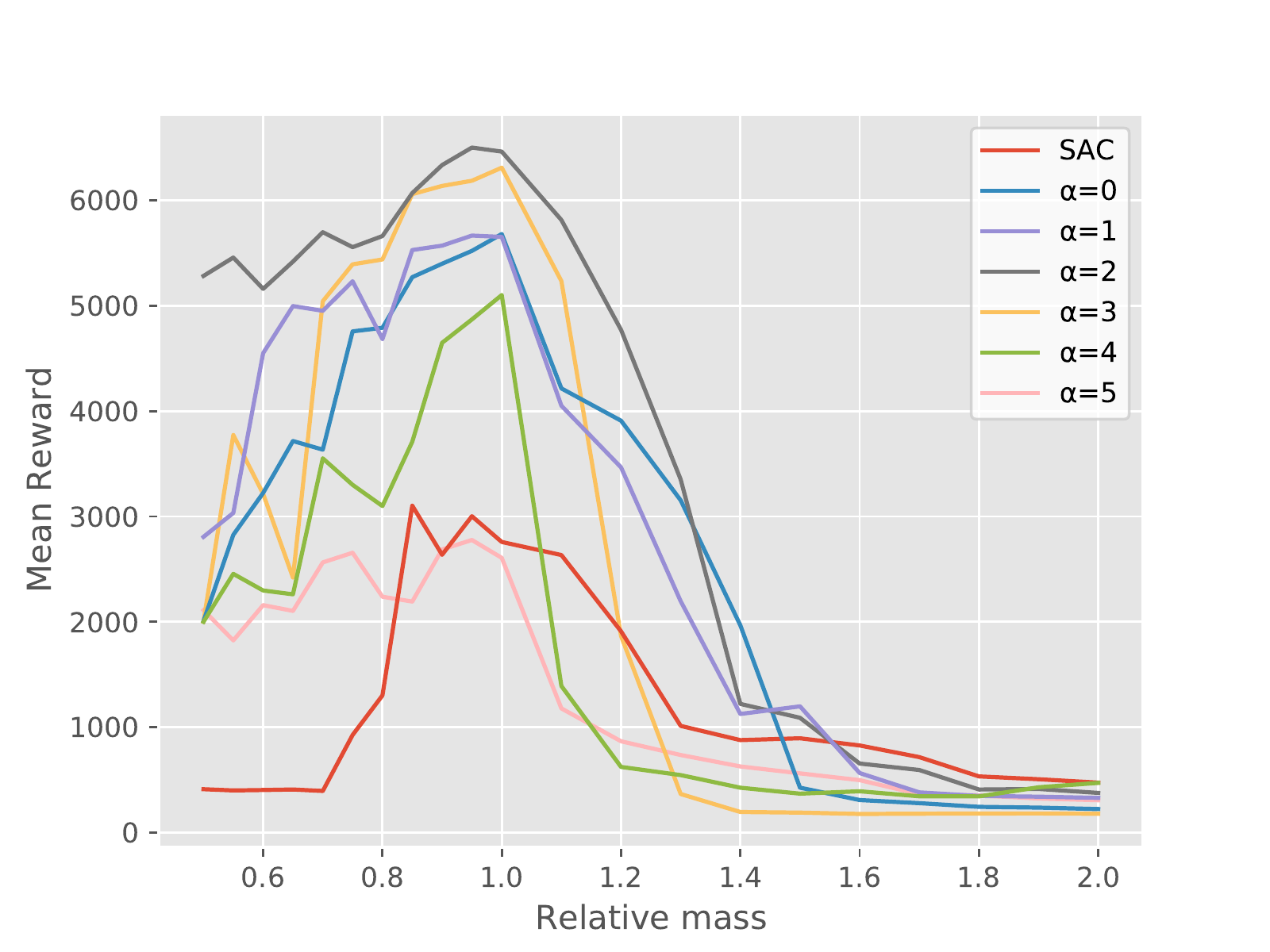}}
      \caption{Mean over 20 trajectories varying relative mass of environments.}
     \label{main_fig}
\end{figure}

In Fig~\ref{main_fig}, we see that we can find a value of $\alpha$ where the robustness is clearly improved without deteriorating the average performance. Normalized performance using by the maximum of the performance for every curve to highlight robustness and not only mean-performance can be found in   annex B.1. If a too strong penalty is applied, the average performance can be decreased as in the HalfCheetah-v3 environment (see annex B.1). For Hopper-v3, a $\alpha$ calibrated at 5 gives very good robustness performances while for Walker2d-v3, the value is closer to 2. This phenomenon was expected and in agreement with our formulation. Moreover, our algorithm outperforms the SAC algorithm for Robustness tasks in all environments. The tuning of $\alpha$ is discussed in annex B.1.

The second surprising observation is that penalizing our objective also improves performance in terms of stability during training and in terms of average performance, especially for Hopper-v3 and Walker2d-v3 in Fig~\ref{main_fig}. Similar results are present in the work of \citep{moskovitz2021tactical} which gives an interpretation in terms of optimism and pessimism for the environments. This phenomenon is not yet explained, but it is present in some environments that are particularly unstable and have a lot of variance. 

We observe similar results for discrete environments such as Cartpole-v1 and Acrobot-v1 (See annex B.2.). 

\vspace{-0.5cm}

\section{Conclusions}

In this paper, we have tried to show that by using a mean-standard deviation formulation to choose our actions pessimistically, we can increase the robustness of our environment for continuous and discrete environments without adding complexity. A single fixed $\alpha$ parameter must be tuned to obtain good performance without penalizing the average performance too much. Moreover, for some environments, it is relevant to penalize to increase the average performance as well when there is lots of variability in the environment.

About the limitations of this work, the convergence of the algorithm to a fixed point is not shown for mean-standard deviation penalization and this formulation is still based on a heuristic even if the link between robustness and penalization is established. Theoretical link with a \citep{kumar2022efficient} could be an interesting way of analyzing our algorithm as they use action value to penalize their objective and remains a way to explain this phenomenon. This is left for future work.

\nocite{*}

\newpage

\bibliography{example_paper}
\bibliographystyle{icml2022}

\newpage
\appendix

\onecolumn

\section{Proof of (\ref{eq_cauchy})}

We consider the following equality :

\begin{equation}
\label{theorem_cauchy}
    \min  _{P \in D_{\chi^2}(P\| P_0)\leq \alpha} Q^{(P,\pi)}= Q^{(P_0,\pi)}-\alpha\mathbb{V}_{P_0}[Z]^{\frac{1}{2}}.
\end{equation}

Consider that trajectories $\tau$ is drawn from $\mathbb{P}$ but here we will write $P$ the transition of the environment as the policy $\pi$ is fixed and it is the only part which differ.

Writing $\tilde{R}(\tau)=R(\tau)-\mathbb{E}_{\tau \sim P_0}[R(\tau)]$ we get :
\begin{align*}
     \norm{ \mathbb{E}_{\tau \sim P}[R(\tau)]-\mathbb{E}_{\tau \sim P_0}[R(\tau)] } &=\norm{  \int_{\tau} \tilde{R}(\tau) \big(p(\tau)-p_0(\tau) \big)d\tau  }\\
     &= \norm{  \int_{\tau}  \tilde{R}(\tau)\sqrt{p_0(\tau)} \frac{\big(p(\tau)-p_0(\tau) \big)}{\sqrt{p_0(\tau)}} d\tau     }\\
     & \leq \norm{ \int_{\tau}   \tilde{R}(\tau)^2 p_0(\tau) d\tau     }^{\frac{1}{2}} \norm{   \int_{\tau} \frac{\big(p(\tau)-p_0(\tau) \big)^2}{p_0(\tau)} d\tau     }^{\frac{1}{2}}\\
     &=\mathbb{V}_{P_0}[R(\tau)]^{\frac{1}{2}}D_{\chi^2}(P\| P_0)^{\frac{1}{2}},
\end{align*}

 because of the positivity of divergence and of the variance, norms are removed. This inequality comes from Cauchy-Schwarz inequality and becomes an equality if for $\lambda \in \mathbb{R}$ :

\begin{equation}
\label{equality}
      \tilde{R}(\tau)p_0(\tau)=\lambda(p(\tau)-p_0(\tau)) \iff p(\tau)=p_0(\tau)(1+\frac{1}{\lambda} \tilde{R}(\tau)).
\end{equation}

However, $ p(\tau)$ needs to be non-negative and sum to one as it is a measure. Normalisation condition is respected by construction however to ensure that the measure is non-negative, this requires $\norm{\tilde{R}(\tau)/\lambda} \leq 1$  in the case where  $\lambda\leq 0$ . In this case of equality, we obtain  from  \ref{equality} that $D_{\chi^2}(P\| P_0)=\frac{\mathbb{V}_{P_0}[R(\tau)]}{\lambda^2}$. Replacing the divergence in the inequality, the following result holds :

\begin{align*}
     \norm{ E_{\tau \sim P}[R(\tau)]-E_{\tau \sim P_0}[R(\tau)] } \leq \frac{\mathbb{V}_{P_0}(R(\tau))}{\lambda}.
\end{align*}

For proving (\ref{theorem_cauchy}) we are interested in the case where  $D_{\chi^2}(P\| P_0)\leq \alpha$, from the initial inequality we obtain :

$$    \min  _{P \in D_{\chi^2}(P\| P_0)\leq \alpha} Q^{(P,\pi)}\geq   \min  _{P \in D_{\chi^2}(P\| P_0)\leq \alpha} Q^{(P_0,\pi)}-D_{\chi^2}(P\| P_0)\mathbb{V}_{P_0}[Z]^{\frac{1}{2}}=Q^{(P_0,\pi)}-\alpha\mathbb{V}_{P_0}[Z]^{\frac{1}{2}}$$

 with the maximum value of $\alpha$ equals to $D_{\chi^2}(P\| P_0)=\frac{\mathbb{V}_{P_0}[R(\tau)]}{\lambda^2}\leq \frac{\mathbb{V}_{P_0}[R(\tau)]}{\norm{\tilde{R}}_{\infty}^2}=\frac{\norm{\tilde{R}}^2_2}{\norm{\tilde{R}}_{\infty}^2}\leq 1$, where the first inequality comes from the conditions $\norm{\tilde{R}(\tau)/\lambda} \leq 1$ and the last one comes from that the $L_2$ norm is smaller than $\infty$-norm.

If our problem is contrained, assuming $\alpha\leq \frac{\mathbb{V}_{P_0}[R(\tau)]}{\norm{\tilde{R}}_{\infty}^2}\leq 1$, we obtain the following results with the maximum attained for $D_{\chi^2}(P\| P_0)= \alpha$ :

\begin{equation}
    \min  _{\mathcal{P} \in D_{\chi^2}(P\| P_0)\leq \alpha} Q^{(\mathcal{P},\pi)}= Q^{(\mathcal{P}_0,\pi)}-\alpha\mathbb{V}_{}[Z]^{\frac{1}{2}}.
\end{equation}

For $\alpha> 1$, we still optimize a lower bound of the quantity of interest. The formulation of our algorithm becomes:

$$
\left\{\begin{array}{l}
\pi_{k+1} \in \mathcal{G}_\alpha\left(Z_{k}\right)=\mathcal{G}(\xi_\alpha(Z_k)=\argmax \limits_{\pi \in \Pi } \langle      \mathbb{E}[Z_k] - \alpha \sqrt{\mathbb{V}[Z_k]}, \pi\rangle  \\
Z_{k+1}=\left(T^{\pi_{k+1}}\right)^{m} Z_{k}
\end{array}.\right.
$$

\section{Further Experimental Details}
All experiements were run on a cluster containing  an Intel Xeon CPU Gold 6230, 20 cores and every single experiments was performed on a single CPU between 3 and 6 hours for continuous control and less than 1 hour for discrete control environment.

Pre-trained models will be available for all algorithms and environments on a GitHub link.

The Mujoco OpenAI Gym tasks licensing information is given at \hyperlink{https://github.com/openai/gym/blob/master/LICENSE.md}{https://github.com/openai/gym/blob/master/LICENSE.md}. Baseline implementation of PPO, SAC, TQC and QRDQN can be found in \citet{raffin2019stable}. Moreover, hyperparameters across all experiments used are displayed in Table \ref{sample-table}, \ref{sample-table2} and \ref{sample-table4} .

\subsection{Results for continuous action spaces}

Tuning of $\alpha$ must be chosen carefully, for example,  $\alpha$ is chosen in $\{0,1,...,5\}$ for  Hopper-v3 and Walker2d-v3 whereas values of $\alpha $ are chosen smaller in $\{0,0.1,0.5.1,1.5,2\}$ and not in a bigger interval. As a rule of thumb for choosing $\alpha$, we can look at the empirical mean and variance at the end of the trajectories to see if the environment has rewards that fluctuate a lot. The smaller the mean/variance ratio, the more likely we are to penalise our environment. For HalfCheetah-v3, the mean/variance ratio is about approximately 100, so we will favour smaller penalties than for Walker2d where the mean/variance ratio is about 50 or 10 for Hopper-v3.

\begin{figure}[h!]
     \centering
     \subfloat[Hopper]{ \includegraphics[width=0.2\textwidth]{final_folder/Hopper_test_non.pdf}}
     \hspace{1cm}
     \subfloat[Walker2d]{ \includegraphics[width=0.2\textwidth]{final_folder/walker_test_non.pdf}}
     \hspace{1cm}
     \subfloat[HalfCheetah]{ \includegraphics[width=0.2\textwidth]{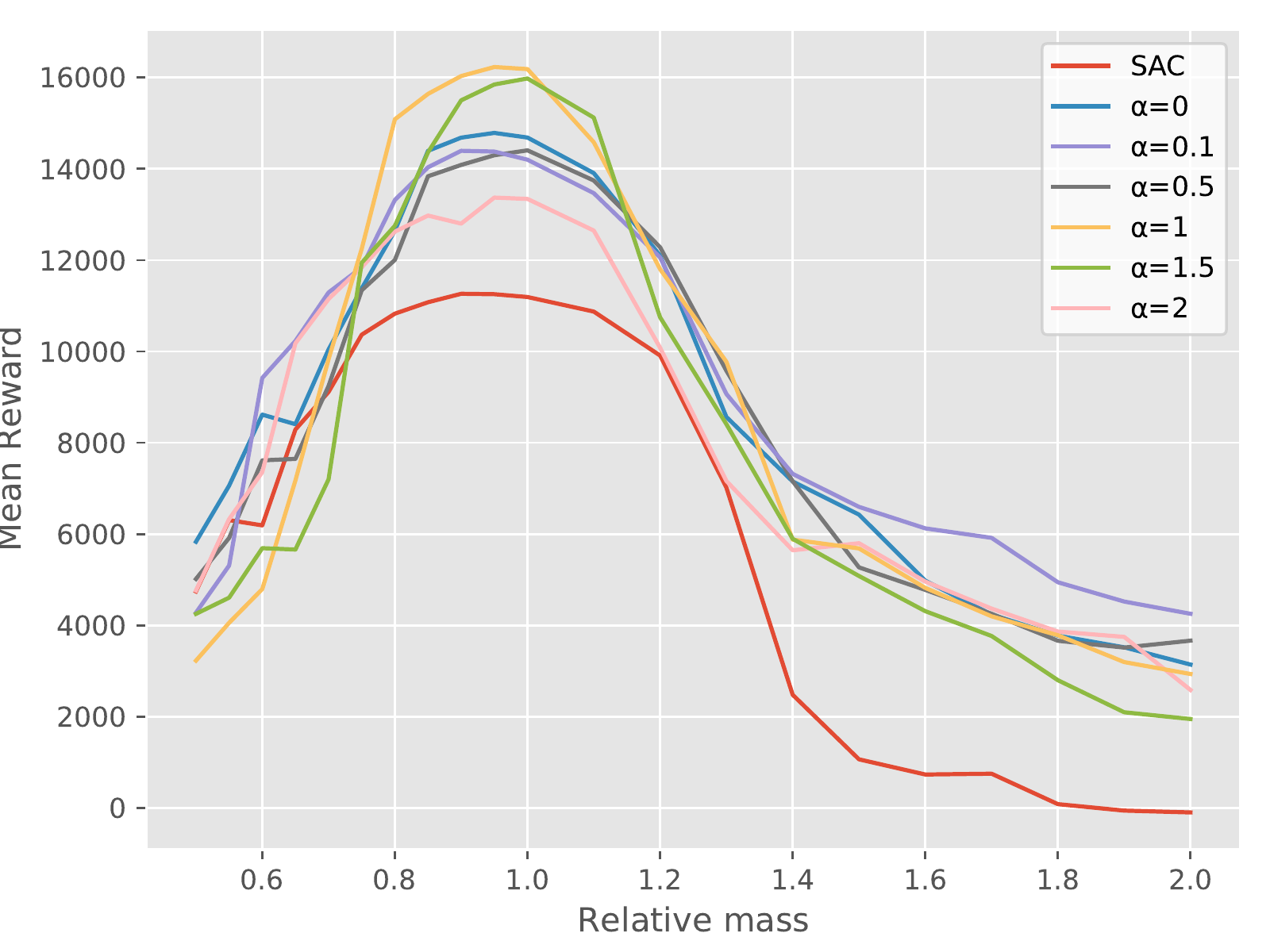}}
      \caption{Mean results on 20 trajectories varying relative mass.}
     \label{steady_state_5}
\end{figure}

\begin{figure}[h!]
     \centering
     \subfloat[Hopper]{ \includegraphics[width=0.2\textwidth]{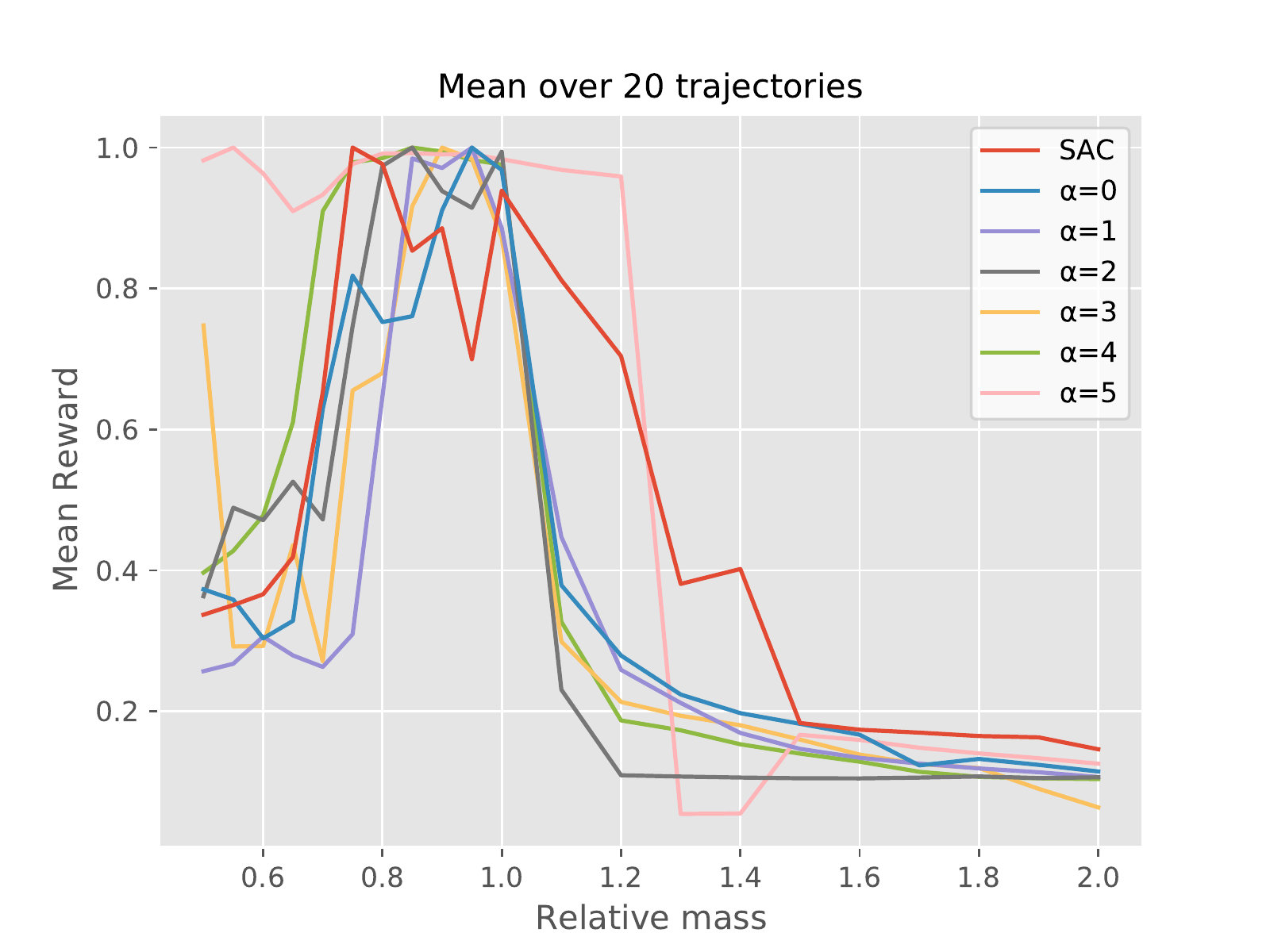}}
     \hspace{1cm}
     \subfloat[Walker2d]{ \includegraphics[width=0.2\textwidth]{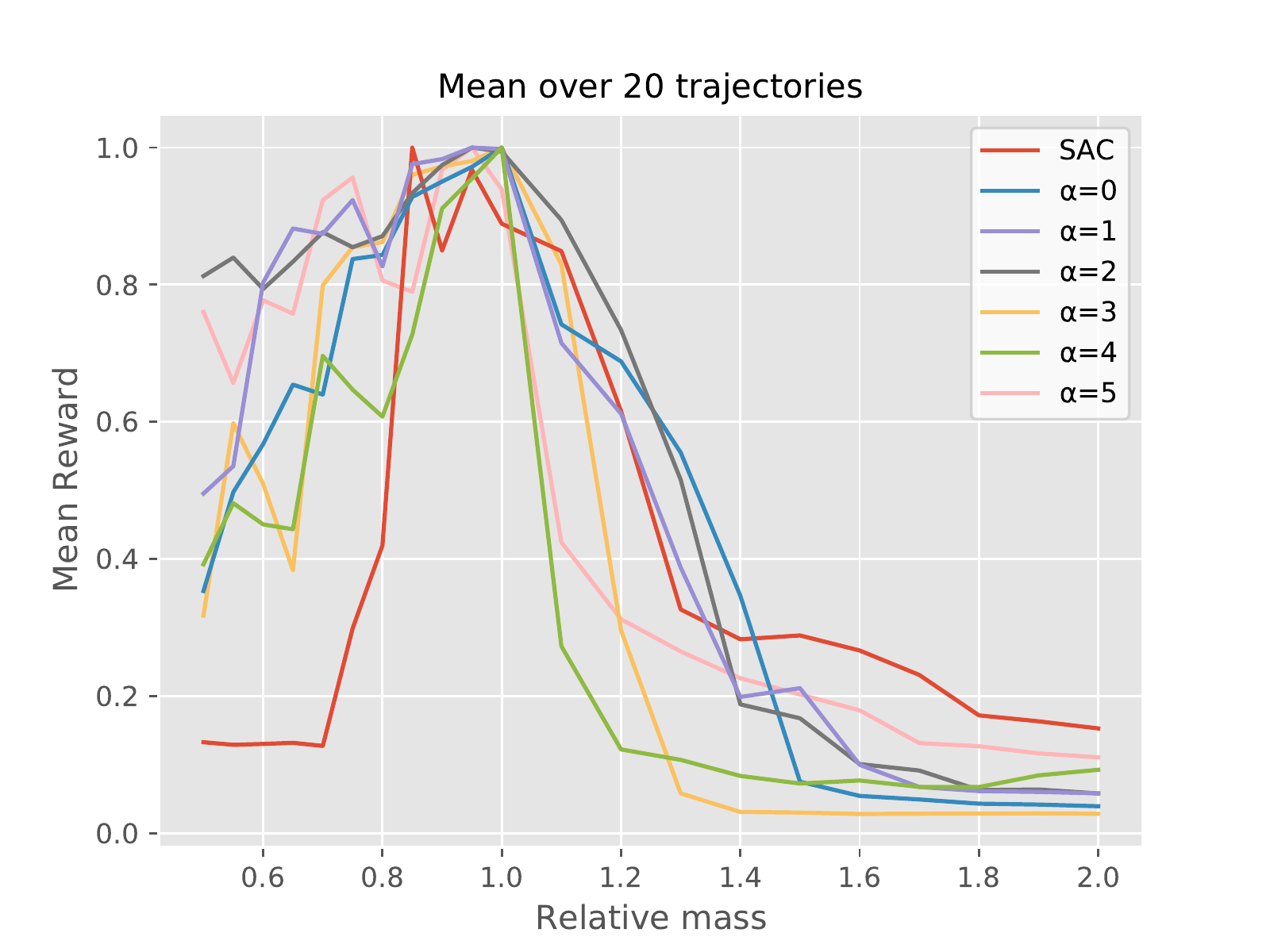}}
     \hspace{1cm}
     \subfloat[HalfCheetah]{ \includegraphics[width=0.2\textwidth]{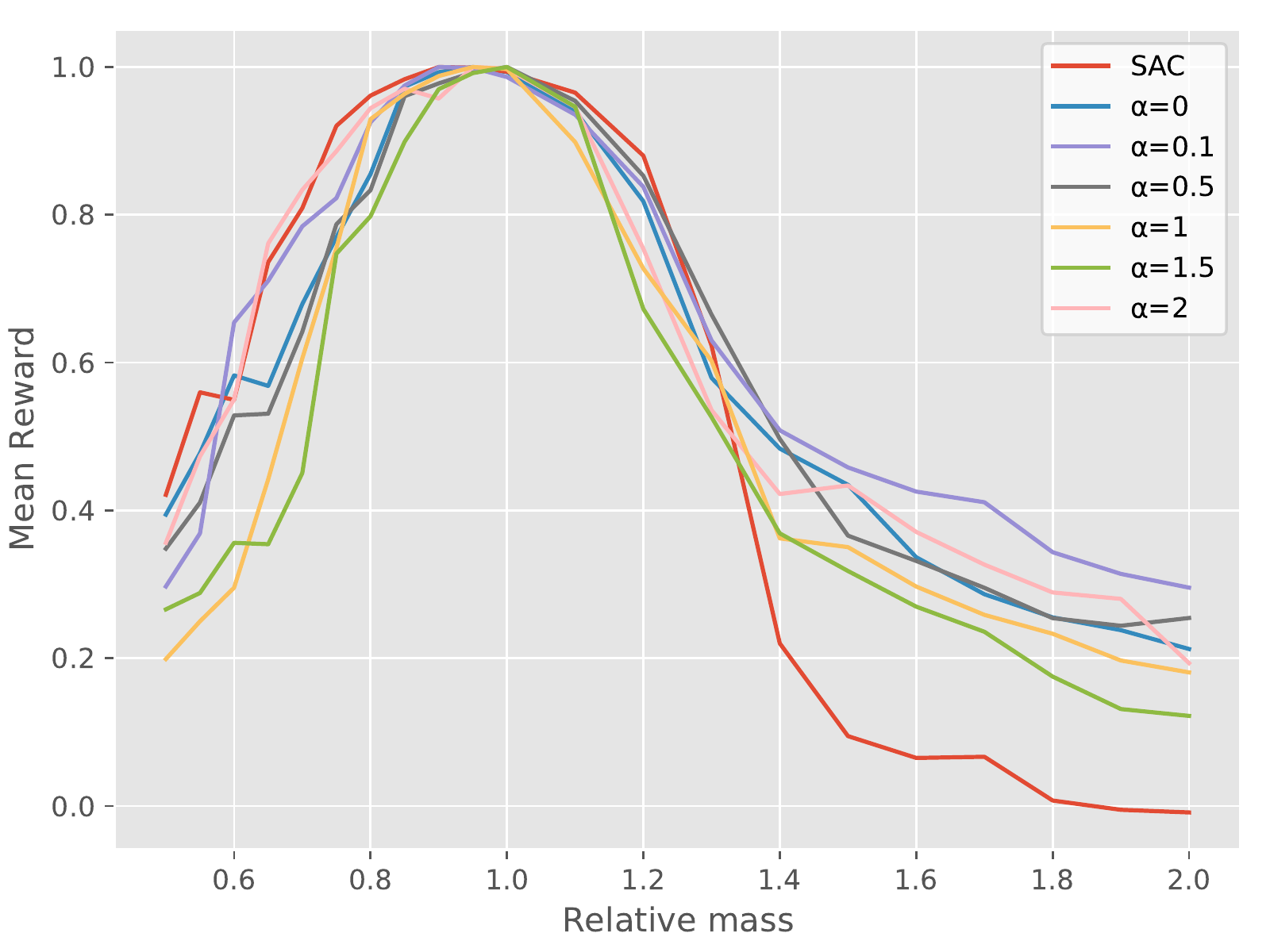}}
      \caption{Normalised mean results on 20 trajectories varying relative mass.}
     \label{steady_state_6}
\end{figure}

\begin{figure}[h!]
     \centering
     \subfloat[Hopper]{ \includegraphics[width=0.2\textwidth]{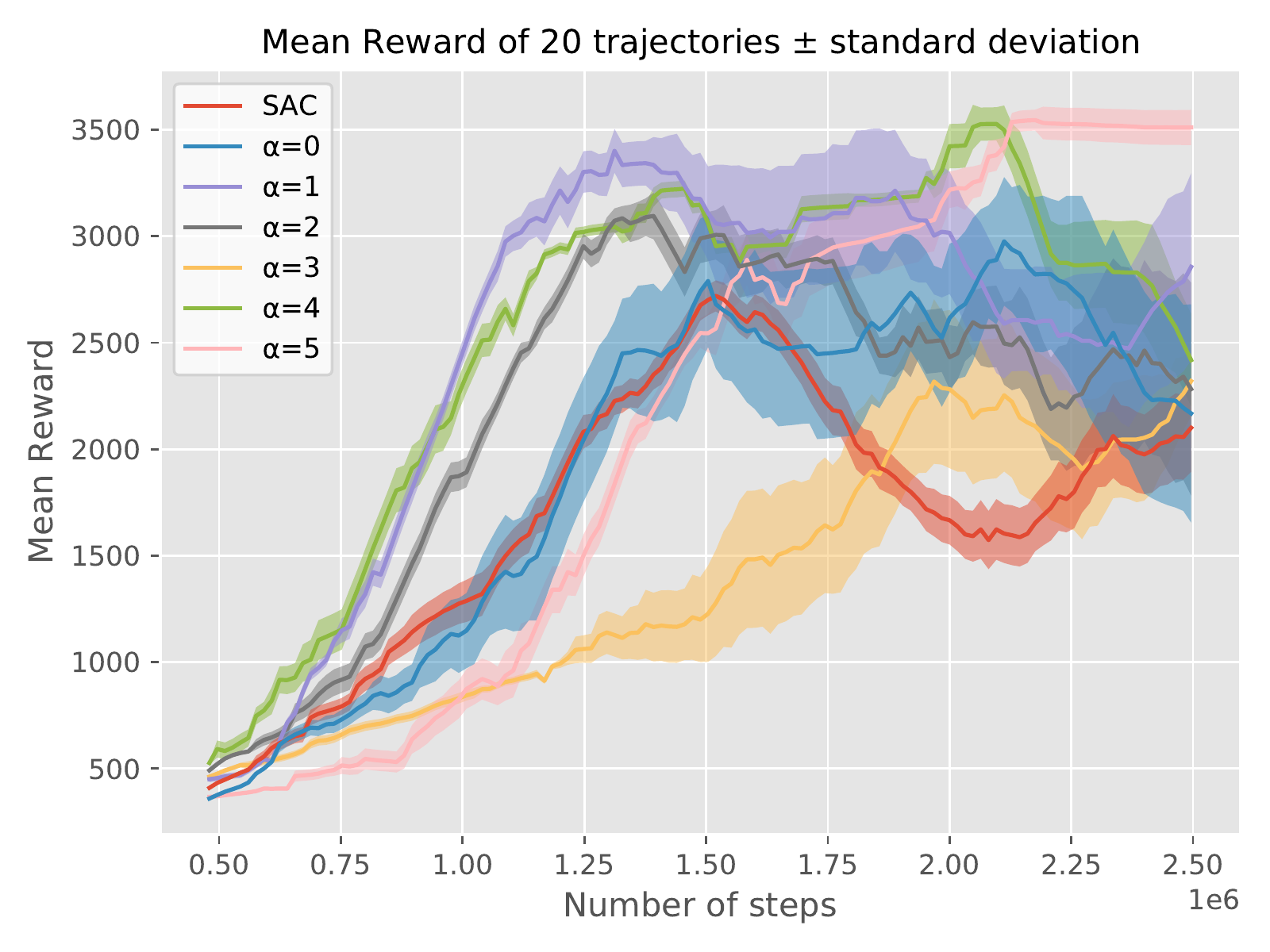}}
     \hspace{1cm}
     \subfloat[Walker2d]{ \includegraphics[width=0.2\textwidth]{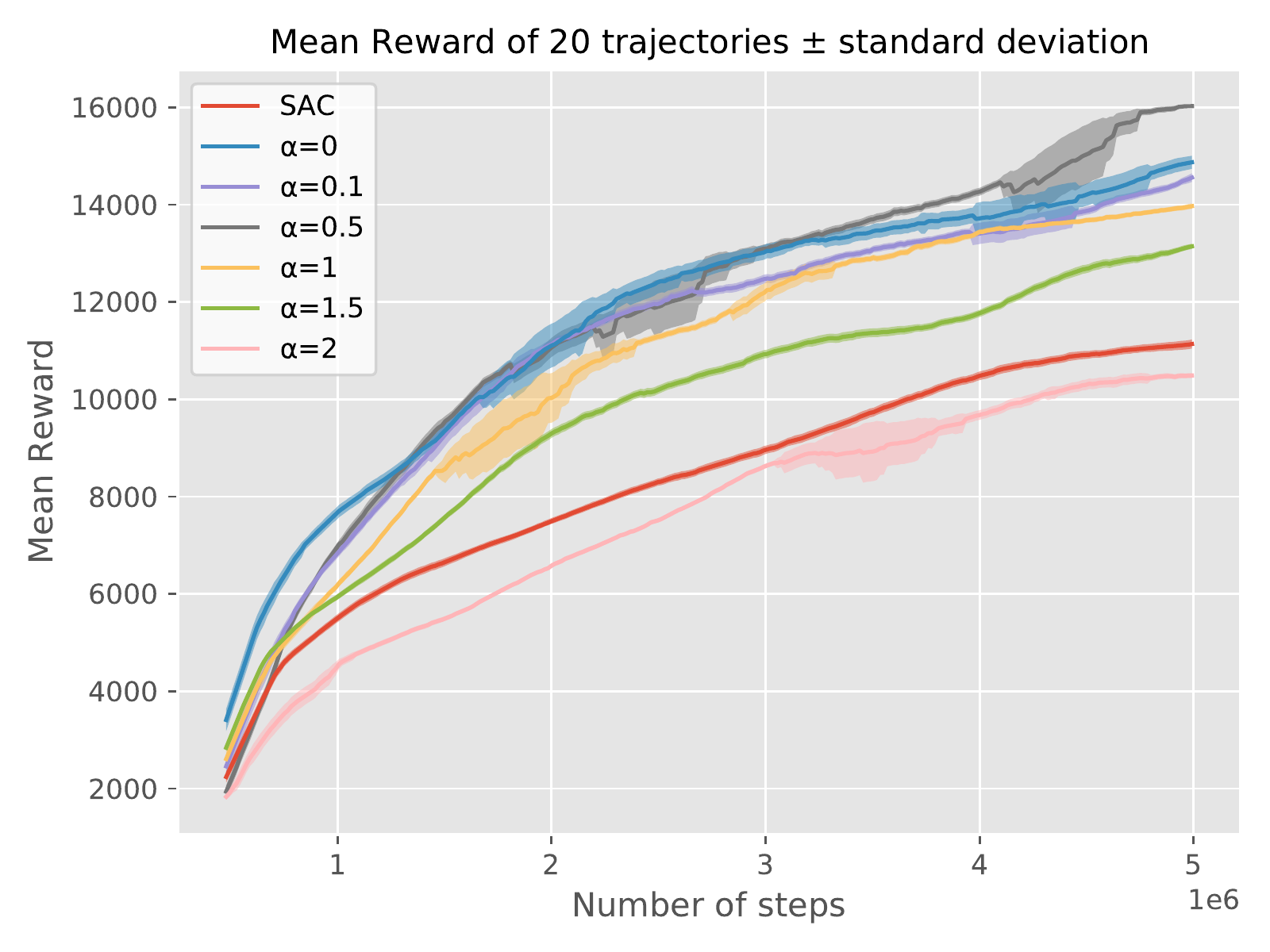}}
     \hspace{1cm}
     \subfloat[HalfCheetah]{ \includegraphics[width=0.2\textwidth]{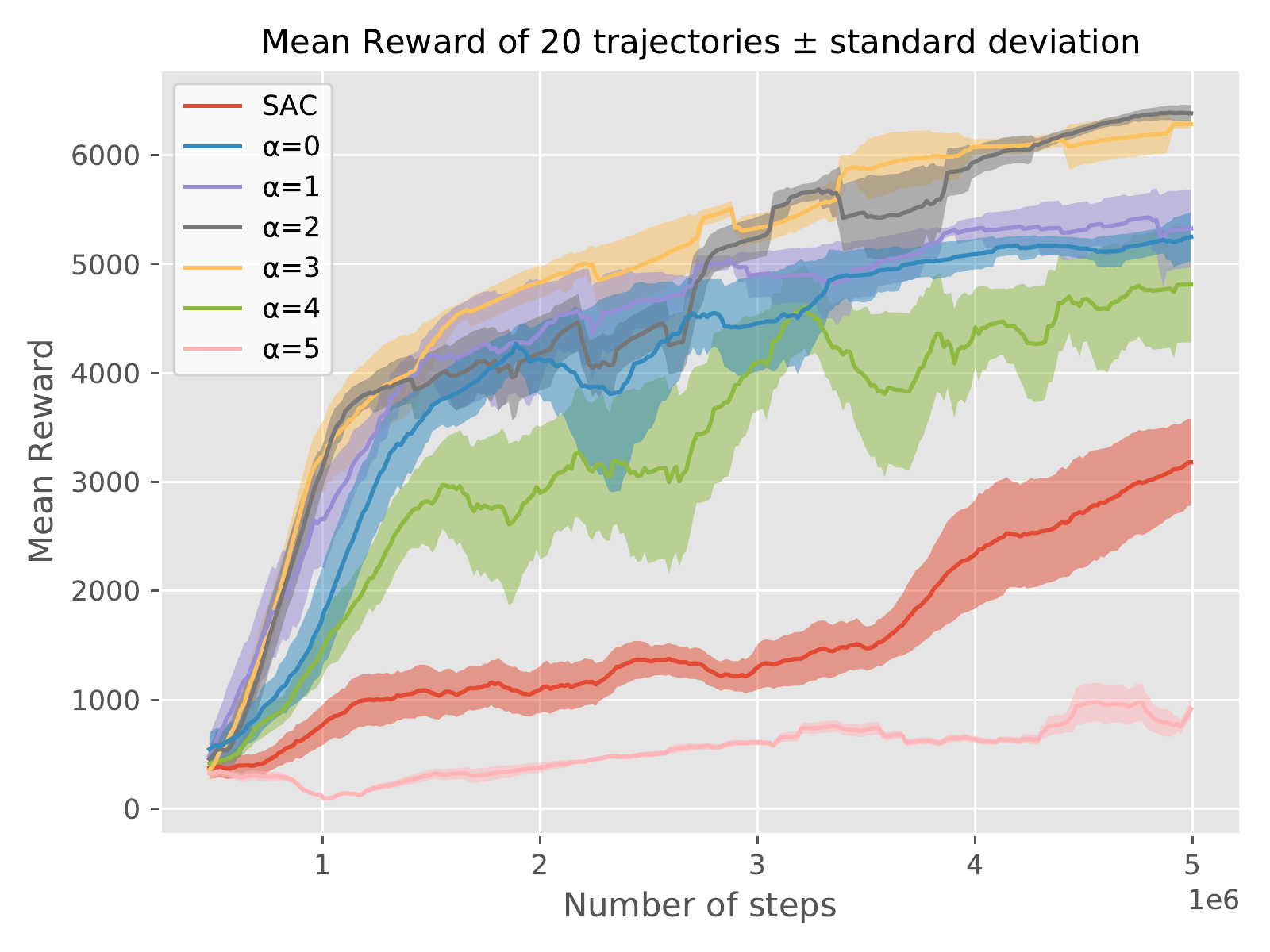}}
     \caption{Mean performances over 20 trajectories ± variance.}
     \label{perf}
\end{figure}

\newpage

\subsection{Results on discrete action spaces}

We test our algorithm QRDQN with standard deviation penalization on discrete action space, varying the length of the pole in Cartpole-v1 and Acrobot-v1 environments. We observe similar results for the discrete environment in terms of robustness. Training is done for a length of the pole equal to the x-axis of the black star on the graph, then for testing, the length of the pole is increased or decreased. We show that robustness is increased when we penalised our distributional critic.  We have compared our algorithm to PPO which has shown relatively good results in terms of robustness for discrete action space in \citep{abdullah2019wasserstein} as SAC does not apply to discrete action space. The same phenomenon is observed in terms of robustness as for the continuous environments. However, the more surprising observations on Hopper and Walker2d with an improvement in performance on average is to be qualified. This is partly explained by the fact that the maximum reward is reached in Cartpole and Acrobot quickly. An ablation study can be found in annex C  where we study the impact of penalization on our behavior policy during testing and on the policy used during learning. It is shown that both are needed in the algorithm.


\begin{figure}[!h]
    \centering
        \includegraphics[width=0.4\textwidth]{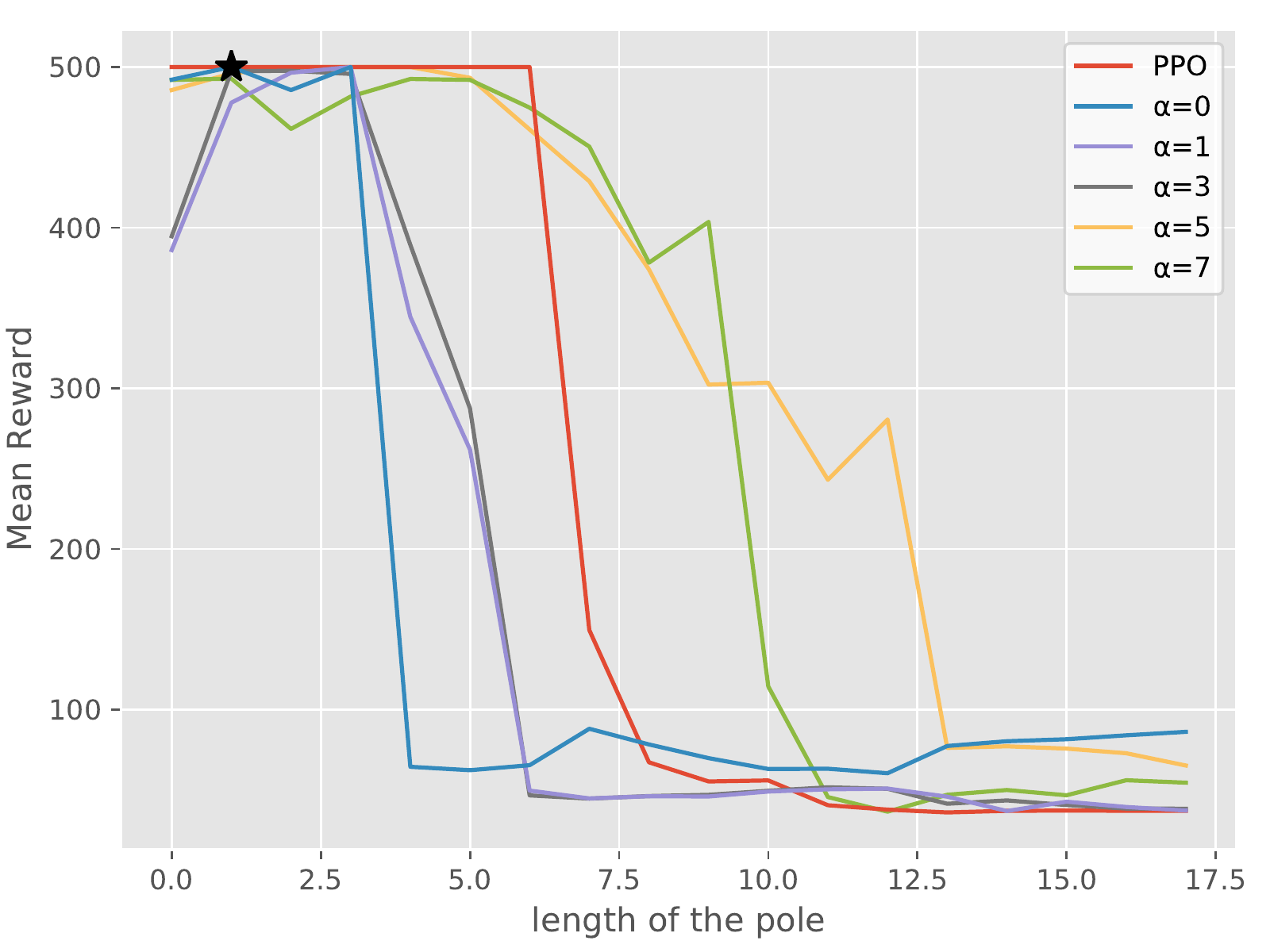}
        \includegraphics[width=0.4\textwidth]{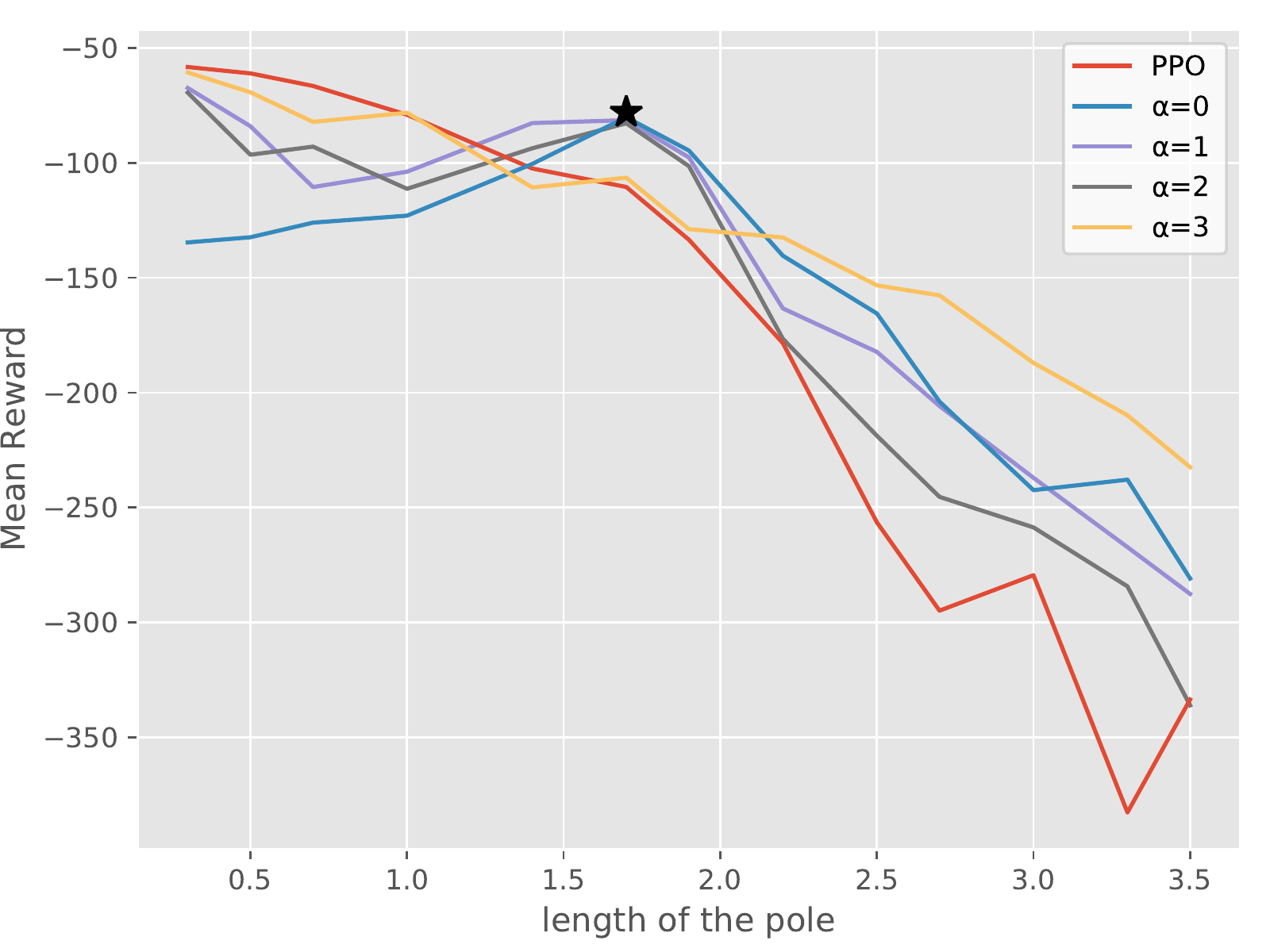}
    \caption{Mean over 20 trajectories varying length of the pole trained on the x-axis of the black star for Cartpole-v1 and Acrobot-v1 environments }
\end{figure}

\subsection{Ablation study}

The purpose of this ablation study is to look at the influence of penalization in the discrete action space with QRDQN. In the figures below, we look at the influence of penalizing only during training, which will have the effect of choosing less risky actions during training in order to increase robustness. This curve is denoted \textit{Train penalized.} 

Then we look at the influence of penalizing only once the policy has been learned using classic QRDQN without penalization. Only mean-var actions are selected here during testing and not during training. This experience is denoted \textit{Train Penalization}.

Finally, we compare its variants with our algorithm called \textit{Full penalization.} The results of the ablation are: to achieve optimal performance, both phases are necessary. 

When penalties are applied only during training. Good performance is obtained in general close to the length 1 where we train our algorithm. However, the performance is difficult to generalize when the pole length is increased as we do not penalize during testing.

When we penalize only during testing: even if the performances deteriorate, we see that it tends to add robustness because the curves have less tendency to decrease when we increase the length of the pole. The performances are not very high as we play different acts than those taken during the learning.

So both phases are therefore necessary for our algorithm. Penalizing during training allows for safer exploration and penalizing during testing allows for better generalization.

The ablation study for the continuous case is more difficult to do. Indeed, the fact that the penalty occurs only in the gradient descent phase makes it difficult to penalize only in the test phase.

\begin{figure}[h!]
     \centering
     \subfloat[$\alpha=1$]{\includegraphics[width=0.25\textwidth]{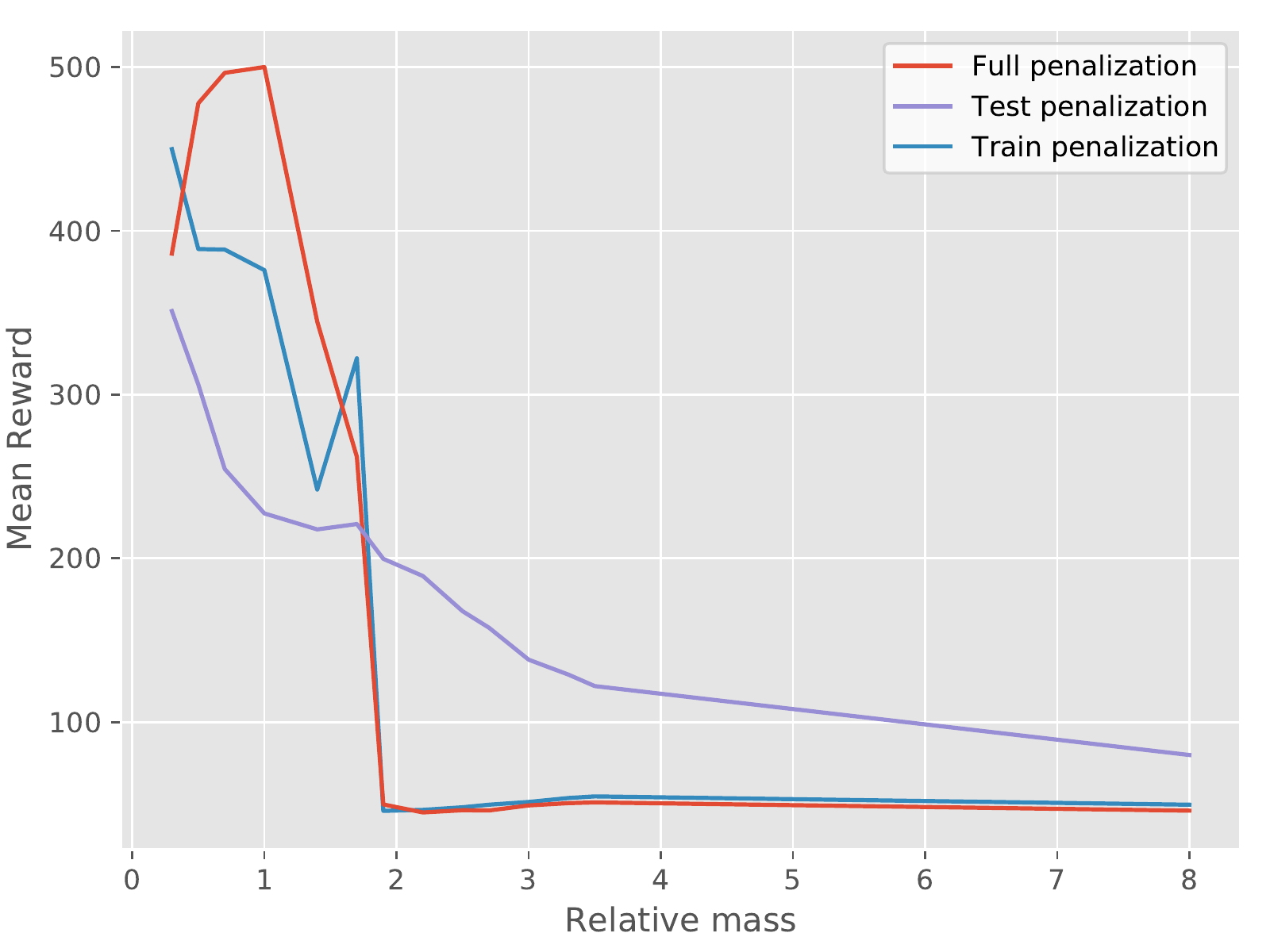}}
     \hspace{2cm}
     \subfloat[$\alpha=3$]{\includegraphics[width=0.25\textwidth]{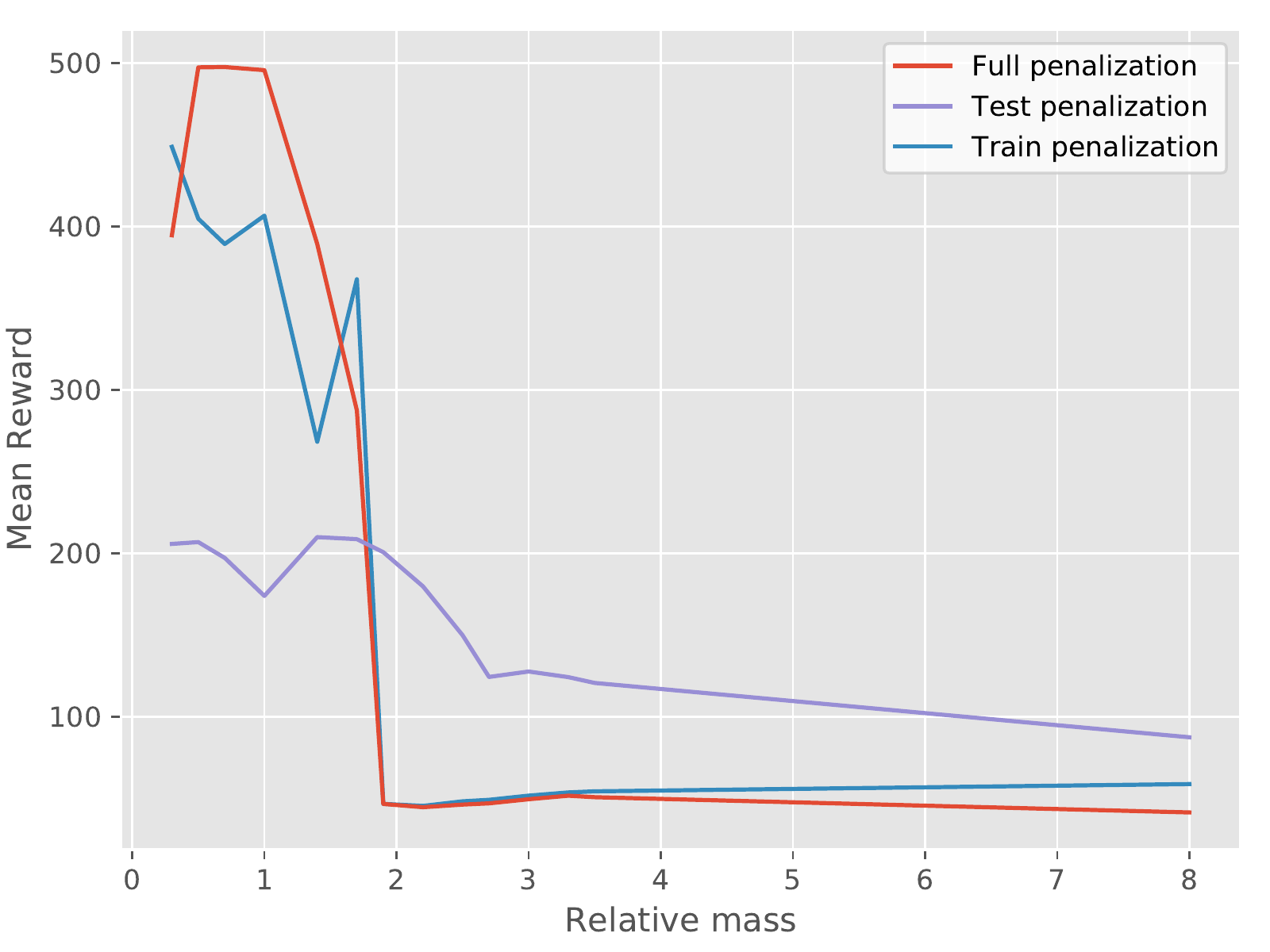}}
     
     \label{steady_state_4}
\end{figure}

\begin{figure}[h!]
     \centering
     \subfloat[$\alpha=1$]{ \includegraphics[width=0.25\textwidth]{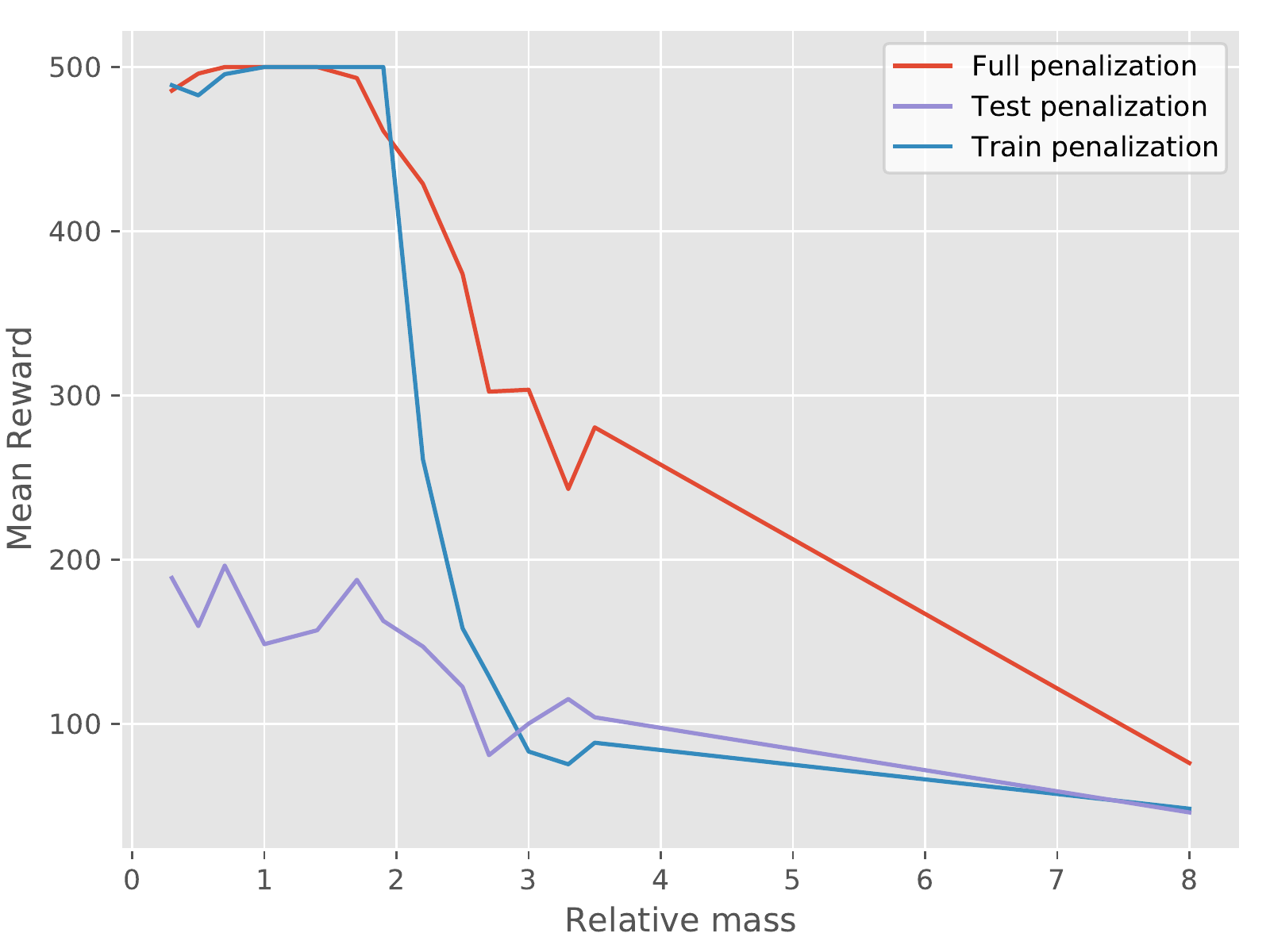}}
     \hspace{2cm}
     \subfloat[$\alpha=3$]{\includegraphics[width=0.25\textwidth]{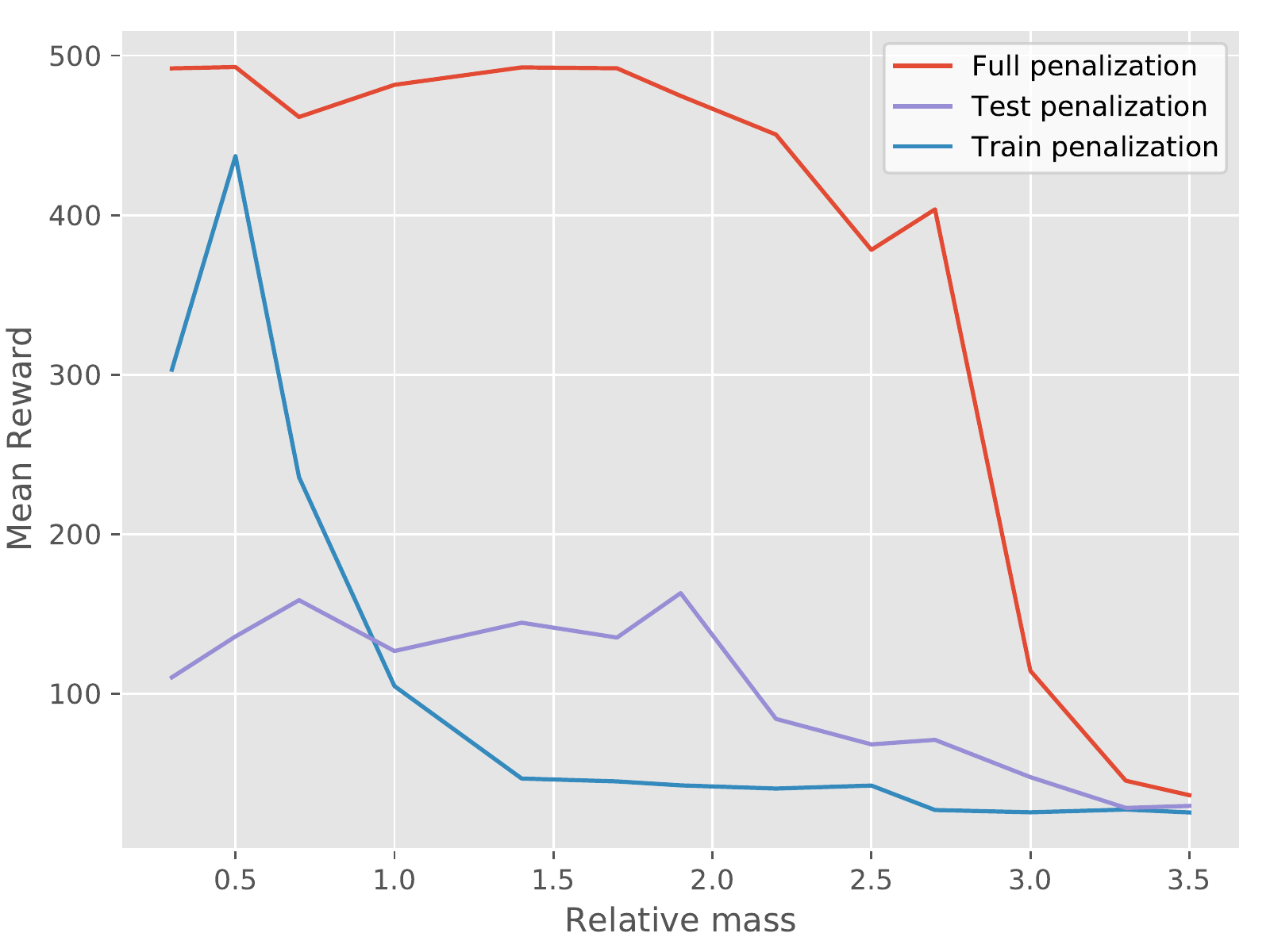}}
     \caption{Ablation study for different values of $\alpha$}
     \label{steady_state_4}
\end{figure}

\newpage

\section{Hyperparameters}

For HalfCheetah-v3 , penalisation is chosen in $[0,2]$ and not $[0,5]$ like in Walker-v3 and Hopper-v3.

\begin{table}[h!]
  \caption{Table of best hyperparameter for Cartpole-v1 }
  \label{sample-table2}
  \centering
  \begin{tabular}{lll}
    
    \multicolumn{2}{c}{}                   \\
    \cmidrule(r){1-3}
    \textbf{Hyperparameter}     & \textbf{QRDQN with standard deviation penalisation}    & \textbf{PPO} \\
    \midrule
   
    Learning Rate    & 2.3e-3 & 3e-4      \\
    Optimizer & Adam & Adam \\
    Replay Buffer Size   & 10e5       &  N/A  \\
    Number of Quantiles&       10       &                       N/A  \\
    Huber parameter $\kappa$   &     1         &N/A  \\
    Penalisation $\alpha$ & \{0,1,3,5,7    \}              & N/A \\
    Network Hidden Layers for Policy        &  N/A            & 256:256  \\
    Network Hidden Layers for Critic& 256:256  &  256:256 \\

    Number of samples per Minibatch &64& 256 \\
    Discount factor $\gamma$ & 0.99 & 0.99 \\
    Target smoothing coefficient $\beta$   &.0.005   &N/A  \\
    Non-linearity &ReLu & ReLu    \\
    Target update interval &10&N/A \\
    Gradient steps per iteration &1&1 \\
    Entropy coefficient & N/A& 0 \\
    GAE $\lambda$& 0.95 & 0.8 \\

    \bottomrule
  \end{tabular}
\end{table}

\begin{table}[h!]
  \caption{Table of best hyperparameter for Acrobot-v1 }
  \label{sample-table}
  \centering
  \begin{tabular}{lll}
    
    \multicolumn{2}{c}{}                   \\
    \cmidrule(r){1-3}
    \textbf{Hyperparameter}     & \textbf{QRDQN  with standard deviation penalisation}    & \textbf{PPO} \\
    \midrule
   
    Learning Rate    & 6.3e-4 & 3e-4      \\
    Optimizer & Adam & Adam \\
    Replay Buffer Size   & 50 000       &  N/A  \\
    Number of Quantiles &       25       &                       N/A  \\
    Huber parameter $\kappa$   &     1         &N/A  \\
    Penalisation $\alpha$ & $\{0,0.5,1,2,3\}$              & N/A \\
    Network Hidden Layers for Critic & 256:256&  256:256 \\
    Network Hidden Layers for Policy & N/A&  256:256 \\
    Number of samples per Minibatch &128& 64 \\
    Discount factor $\gamma$ & 0.99 & 0.99 \\
    Target smoothing coefficient $\beta$   &.0.005   &N/A  \\
    Non-linearity &ReLu & ReLu    \\
    Target update interval & 250  & N/A \\
    Gradient steps per iteration &4&1 \\
    Entropy coefficient & N/A& 0 \\
    GAE $\lambda$ &0.95 & 0.95\\

    \bottomrule
  \end{tabular}
\end{table}

\begin{table}[h!]
  \caption{Table of best hyperparameter for all continuous environments }
  \label{sample-table4}
  \centering
  \begin{tabular}{lll}
    
    \multicolumn{2}{c}{}                   \\
    \cmidrule(r){1-3}
    \textbf{Hyperparameter}     & \textbf{TQC with standard deviation penalisation}    & \textbf{SAC} \\
    \midrule
   
    Learning Rate    & linear decay from 7.3e-4 & linear decay from 7.3e-4      \\
    Optimizer & Adam & Adam \\
    Replay Buffer Size   & $10^6$       &  $10^6$  \\
    Expected Entropy Target &  $-\mathrm{dim}\mathcal{A}$            &$-\mathrm{dim}\mathcal{A}$  \\
    Number of Quantiles&       25       &                       N/A  \\
    Huber parameter $\kappa$   &     1         &N/A  \\
    Penalisation $\alpha$ & $\{0,1,...5\}$              & N/A \\
    Network Hidden Layers for Policy        &  256:256            & 256:256  \\
    Network Hidden Layers for Critic& 512:512:512&  256:256 \\

    Number of dropped atoms &2 &N/A \\
    Number of samples per Minibatch &256& 256 \\
    Discount factor $\gamma$ & 0.99 & 0.99 \\
    Target smoothing coefficient $\beta$   &.0.005   &0.005  \\
    Non-linearity &ReLu & ReLu    \\
    Target update interval &1&1 \\
    Gradient steps per iteration &1&1 \\

    \bottomrule
  \end{tabular}
\end{table}


\end{document}